\documentclass[sigconf]{acmart}

\newcommand{\mytag}[1]{{\bf#1}}
\AtBeginDocument{%
  \providecommand\BibTeX{{%
    \normalfont B\kern-0.5em{\scshape i\kern-0.25em b}\kern-0.8em\TeX}}}

\setcopyright{none}
\renewcommand\footnotetextcopyrightpermission[1]{} 
\begin{document}

\title{Visualizing Deep Graph Generative Models for Drug Discovery}


\author{Karan Yang}
\affiliation{%
  \institution{Cornell Tech
}
 \city{New York}
  \state{New York City}
  \country{USA}
}
\email{ky393@cornell.edu}
\author{Chengxi Zang}
\affiliation{%
  \institution{Weill Cornell Medicine 
}
\city{New York}
  \state{New York City}
  \country{USA}
}
\email{chz4001@med.cornell.edu}

\author{Fei Wang}
\affiliation{%
  \institution{ Weill Cornell Medicine  
}
  \city{New York City}
  \state{New York}
  \country{USA}
}
\email{few2001@med.cornell.edu}


\begin{abstract}
 Drug discovery aims at designing novel molecules with specific desired properties for clinical trials. Over past decades, drug discovery and development have been a costly and time-consuming process.  Driven by big chemical data and AI,  deep generative models show great potential to accelerate the  drug discovery process.
 Existing works investigate different deep generative frameworks for molecular generation, however, 
 less attention has been paid to the visualization tools to quickly demo and evaluate model’s results. Here, we propose a visualization framework 
 which provides interactive visualization tools to visualize molecules generated during the encoding-and-decoding process of deep graph generative models, and provide real-time molecular optimization functionalities.
 Our work tries to empower black-box AI-driven drug discovery models with some visual interpretabilities.
 %
 

\end{abstract}




\maketitle
\pagestyle{plain} 

\section{Introduction}
\begin{figure}[!t]
    \centering
    
    \includegraphics[width=0.5\textwidth]{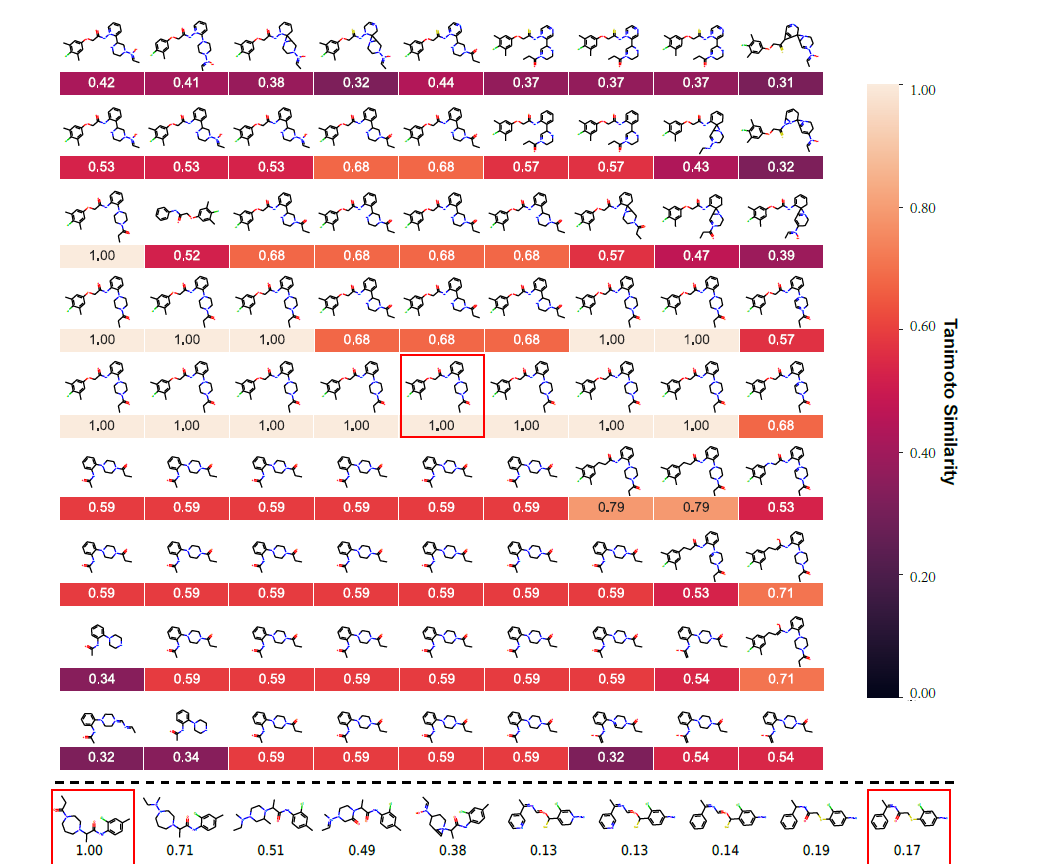}
    \label{fig:VAE}
    \caption{ Visualization of learned latent space by MoFlow\cite{moflow}. Top: Visualization of the grid neighbors of a seed molecule in
the center, which serves as the baseline for measuring similarity. Bottom: Interpolation between two seed molecules and the
left one is the baseline molecule for measuring similarity. Seed molecules are highlighted in red boxs and they are randomly
selected from ZINC250K.\label{fig:latent}  }
\end{figure}

Deep graph generative models are promising to accelerate the long and costly drug discovery process by exploring large chemical space in a data-driven manner. 
These models first learn a continuous latent space by encoding molecular graphs and then generate novel and optimized molecules by decoding from the learned latent space guided by some targeted properties \cite{gomez2018automatic,jin2018junction,moflow}. We illustrate these encoding-and-decoding pipelines in Figure~\ref{fig:latent}. 

Despite their promising results, deep graph generative models are challenging to work with and the training process and results lack transparency and interpretability, in general. These challenges naturally lead us to search for a visualization tool that will help to interpret each step of how the model generates novel chemical structures in its latent space, to decompose the analysis of how the interpolation between molecules are formed and to gain insights over how the optimized functional compounds are yielded. Furthermore, a three-dimensional and interactive molecule visualization tool is needed to develop useful insights - as it is very important and informative to visually examine how the generated molecule’s atoms are positioned relative to each other in 3D space.

\begin{figure}[ht]
    \centering
    \includegraphics[width=0.4\textwidth]{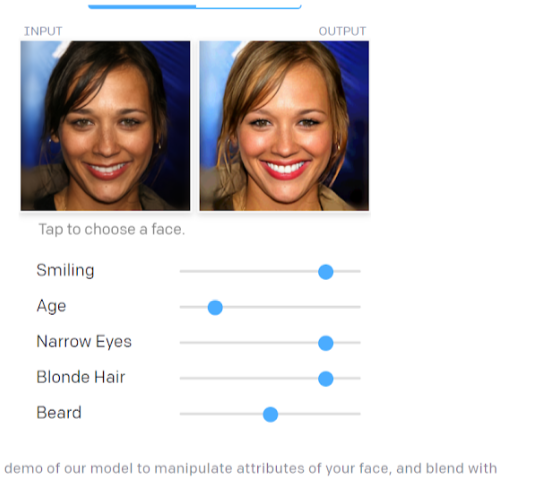}
    \caption{ An interactive demo of the Glow model to manipulate attributes of the face, and blend with other faces.\cite{glow-face} \label{fig:face}} 
\end{figure}

While there are some good visualization tools designed for the deep generative models - such as, “Glow visualization demo”  in Figure~\ref{fig:face}, we have not seen effective visualization tools tailored to the deep generative models for the molecular graph.







In this paper, we propose to build a web dashboard that provides three-dimensional and interactive molecular visualization tools designed specifically for the deep generative models for the molecular graphs. Our dashboard proposes an interface to operate on chemical structures directly with following functionalities:
\begin{itemize}
\item Visualize the generated molecules with different levels of atom outlines, ambient occlusion, and more with the various viewing parameters( angles, rotation, zoom, highlights).  
\item Visualize the chemical similarity between each neighboring molecule and the centering molecule. 
\item Visualize the linear interpolation between
two molecules to show their changing trajectory in the latent space.
\item Visualize the optimized molecules under different objective properties.
\item Query  generated novel molecules based on seed molecules 
at real-time with the deep graph generative model running backend. 
\end{itemize}
Our dashboard works seamlessly with Python-based open source chemoinformatics and machine learning packages such as RDKit \footnote{https://github.com/rdkit/rdkit} and PyTorch/TensorFlow. 


\section{Related Work.}
Interactive interfaces and visualizations of the sampled data from the chemical latent space of deep learning models have been designed and developed to help people understand what models have learned and how they make predictions. Many of those visualization tools require significant workarounds to pre-existing graph types. Here, We discuss three main visualization tools for the  deep graph generative models.

\mytag{RDKit-Neo4j project.} This is a development of extension for neo4j graph database for querying knowledge graphs storing molecular and chemical information\cite{neo4j}. The project’s task is to enable identification of entry points into the graph via exact or substructure/similarity searches. The intention is to use chemical structures as limiting conditions in graph traversals originating from different entry points. Neo4j is coded in CYPHER which is quite complex.  

\mytag{Tableau 3D project.} In a tableau project\cite{tableu}, the Caffeine molecule example uses a dual axis chart. One axis draws the atoms while the other axis draws the bonds between them.  The project relies purely on the background image, mark size and z-order to achieve the 3D look. One issue with Tableau evolves from its dual axis structure, as z-order does not  always work well - as Tableau sorts within each axis independently, so in this setup atoms are always drawn above bonds but each group is sorted within themselves. 

\mytag{“Glow visualization”.}  One interesting online visualization tool for the “Glow”, a reversible generative model, is shown in the figure 2\cite{glow-face}.

However, a limitation of the “Glow visualization” is that it is built upon javascript and only works on the images. Our final goal would be to build an alternative, but similar, interactive visualization tool that is designed specifically for drug discovery purposes and for the generation of the molecular structure under varying parameters.

\section{Method} 
\subsection{MOFLOW MODEL-A Deep Generative Model of Molecular Graphs}

As an initial experiment to build our broad web dashboard, we have underway a process to build a web dashboard specifically for the MoFlow model for molecular graph generation proposed by Zang and Wang\cite{moflow}.  MoFlow is one of the first flow-based models which not only generates molecular graphs at one-shot by invertible mappings but also has a validity guarantee. MoFlow consists of a variant of Glow model for bonds, a novel graph conditional flow for atoms given bonds, and then combining them with
post-hoc validity correction.  MoFlow achieves many new state-of-the-art performance on molecular generation, reconstruction and optimization. We summarize the inference and generation procedure of the MoFlow in Algorithm 1 and Algorithm 2 respectively. 
\begin{figure}[ht]
    \centering
    \includegraphics[width=0.5\textwidth]{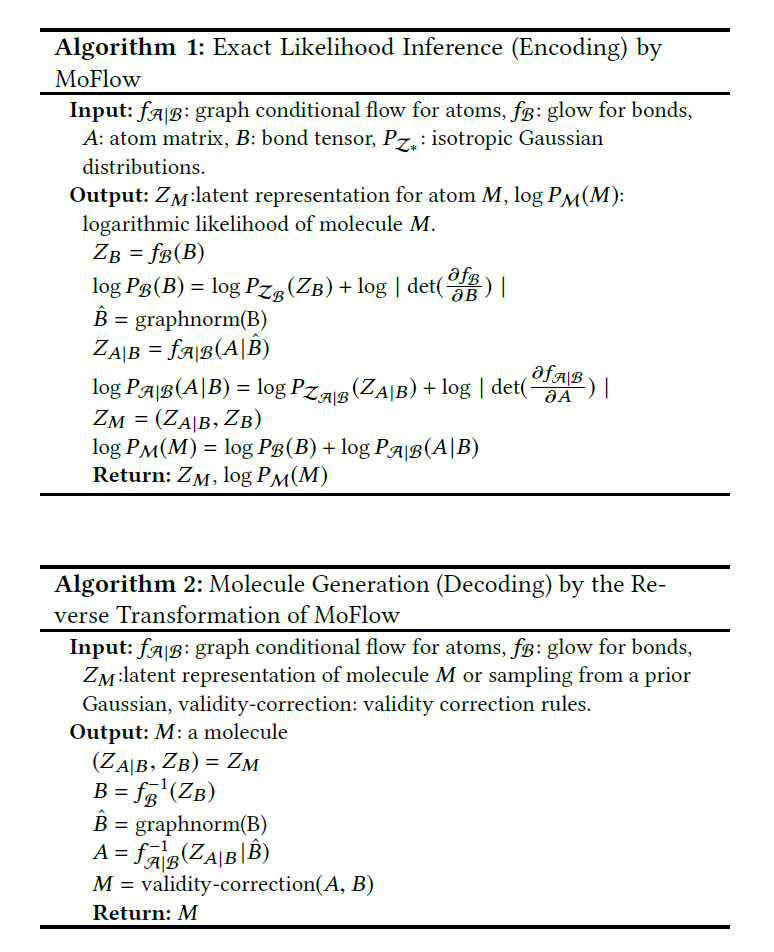}
    
    \label{fig:face2}
\end{figure}

\subsection{Visualization Implementation}

Our approach is to integrate\textbf{ Pytorch, rdkit and Dash/Dash bio} to build a web dashboard that provides three-dimensional and interactive molecule visualization tools. This framework would be able to combine pytorch - an open source machine learning framework, Rdkit - an Open-Source Cheminformatics Software, and Dash - a Python framework for building web applications. Written on top of Flask, Plotly.js, and React.js, Dash is ideal for building data visualization apps with highly customized user interfaces in pure Python. To summarize, the main components of our visualization tools for the molecular generation are:

\begin{itemize}
\item Use Dash as the web dashboard UI interface design. .
\item Callback functions that take an dataset and the best model's parameters as input and connect to the model's encoder and decoder(Algorithm 1 and 2 in figure..)
\item Use Rdkit functions to convert molecular data to 2D view and to “xyz” file for the 3D view.

\item Use Dash Bio-speck\cite{speck} to display the output generated by the model's decoder in an interactive Molecule 3D View 

\end{itemize}

\section{Results}
 We built an efficient web dashboard with a dropdown menu, allowing people to select a drug’s name and display the 2D and 3D interactive view. This dashboard allows users to manipulate attributes to explore the latent space (Note: we can rotate the 3D molecule). Figure 3 and figure 4 are sample demos of our dashboard designs and functionalities.

\begin{figure}[ht]
    \centering
    \includegraphics[width=0.5\textwidth]{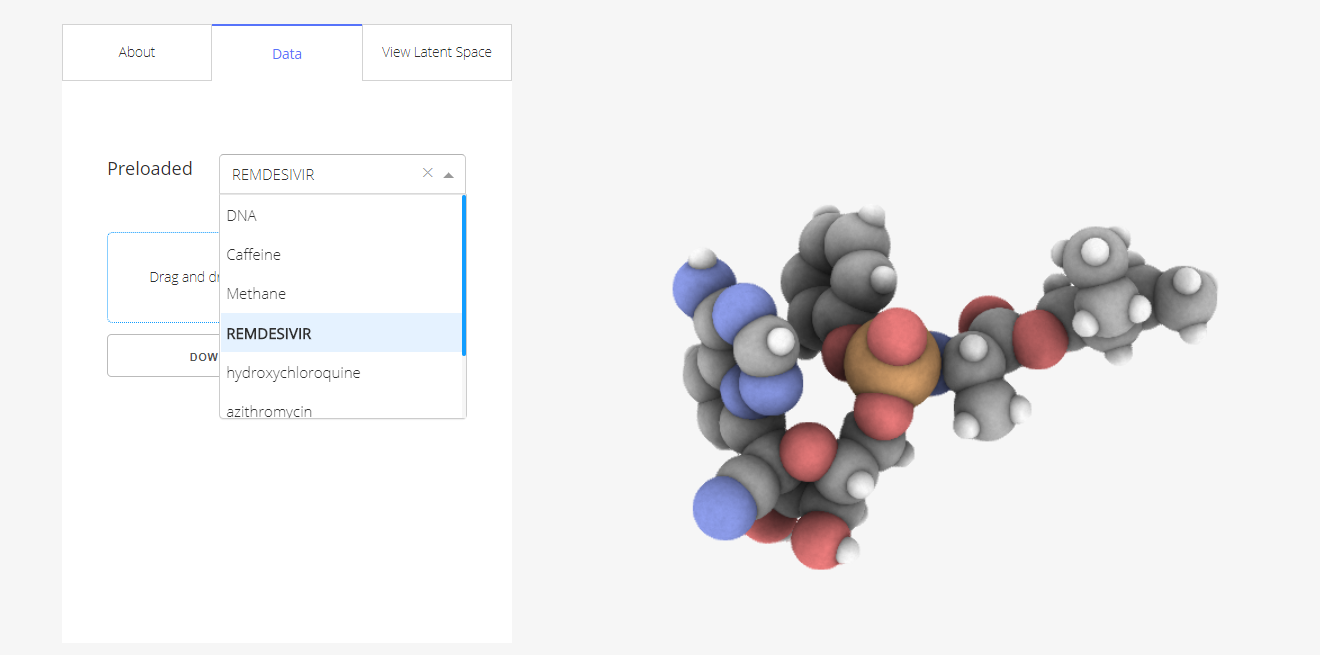}

    \label{fig:dash1}
    \caption{ Demo of 3D interactive visualizations for a specific molecule  }
\end{figure}

\begin{figure}[ht]
    \centering
    \includegraphics[width=0.5\textwidth]{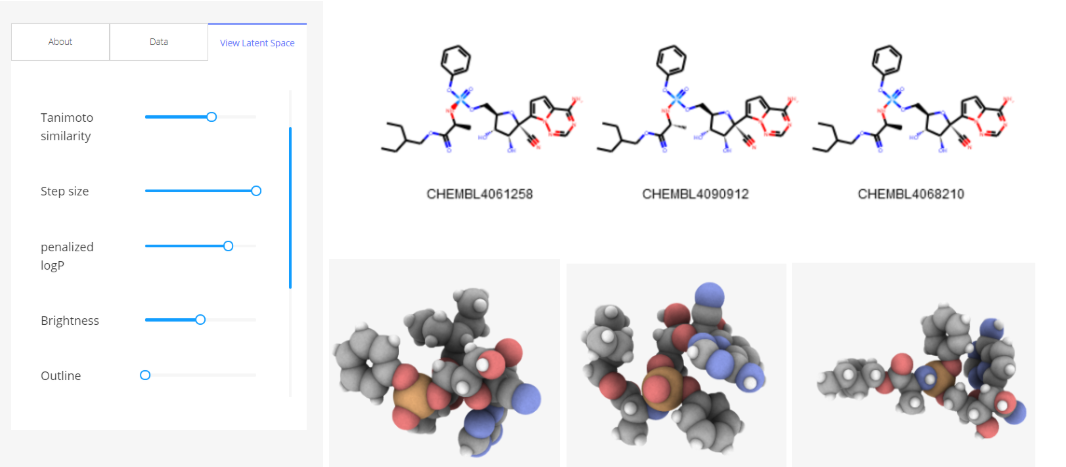}

    \label{fig:dash2}
    \caption{Demo of 2D and 3D interactive visualizations for similar chemical structures in the latent space  }
\end{figure}

\section{Discussion}
We found our approach - which uses Dash / Dash bio - to be ideal for the drug discovery purpose. First, it ties modern UI elements like dropdowns, sliders, and graphs directly to the analytical Python code. Dash isn't just for dashboards. Users will have full control over the look and feel of the applications. The 3D view feature from Dash bio can illustrate the shapes of proteins and provide insights into the way that they bind to other molecules. We can change the level of atom outlines, ambient occlusion, and more with the various viewing parameters. Further, our approach allows for the scroll wheel to control zoom for the molecule. Dash fires Python callback functions when you click on an atom, rotate the molecule, or change the structure. Also, you can highlight individual atoms (like a protein’s active site).\cite{dashbio} These properties render our approach especially valuable when communicating the mechanics of biomolecular processes.

\section{Limitations and Future work}

Currently, our framework only supports specific data file formats (namely, PDB and xyz files). It requires an extra step using Rdkit to convert other file formats to an xyz file. Additionally, most of the components require JSON data as input yet, this file format is not typically provided in datasets or studies.

Our next step is to develop functions to connect deep learning models written in Pytorch at the backend and then output the visualizations to the web dashboard using Dash. The User interface would show the molecular generation process and the optimization process in real-time. Also, future developments to the package therefore should include processing for other important file formats.
\section{Conclusion}
Deep generative models have demonstrated strong potentials on efficient and effective drug molecule design with desired properties.  However, such complex deep learning models for drug discovery are hard to train and hard to understand. In order to provide meaningful solutions to the drug discovery challenge, it is important to have robust data visualization tools for the generative models. Our approach holds the promise for researchers to explore and build on the deep learning models’ results (with less effort) and thereby enable them to focus on evaluation, error analysis and developing valuable insights from the visualizations.  
\small{
\bibliographystyle{ACM-Reference-Format}
\bibliography{main}


\begin{thebibliography}{8}


\ifx \showCODEN    \undefined \def \showCODEN     #1{\unskip}     \fi
\ifx \showDOI      \undefined \def \showDOI       #1{#1}\fi
\ifx \showISBNx    \undefined \def \showISBNx     #1{\unskip}     \fi
\ifx \showISBNxiii \undefined \def \showISBNxiii  #1{\unskip}     \fi
\ifx \showISSN     \undefined \def \showISSN      #1{\unskip}     \fi
\ifx \showLCCN     \undefined \def \showLCCN      #1{\unskip}     \fi
\ifx \shownote     \undefined \def \shownote      #1{#1}          \fi
\ifx \showarticletitle \undefined \def \showarticletitle #1{#1}   \fi
\ifx \showURL      \undefined \def \showURL       {\relax}        \fi
\providecommand\bibfield[2]{#2}
\providecommand\bibinfo[2]{#2}
\providecommand\natexlab[1]{#1}
\providecommand\showeprint[2][]{arXiv:#2}

\bibitem[\protect\citeauthoryear{??}{spe}{[n.d.]}]%
        {speck}
 \bibinfo{year}{[n.d.]}\natexlab{}.
\newblock \bibinfo{title}{Speck Examples and Reference}.
\newblock
\newblock
\urldef\tempurl%
\url{https://dash.plotly.com/dash-bio/speck}
\showURL{%
\tempurl}


\bibitem[\protect\citeauthoryear{BERAN}{BERAN}{2015}]%
        {tableu}
\bibfield{author}{\bibinfo{person}{BORA BERAN}.}
  \bibinfo{year}{2015}\natexlab{}.
\newblock \bibinfo{booktitle}{\emph{Going 3D with Tableau}}.
\newblock
\urldef\tempurl%
\url{https://boraberan.wordpress.com/2015/12/18/going-3d-with-tableau/}
\showURL{%
\tempurl}


\bibitem[\protect\citeauthoryear{Dhariwal and Kingma}{Dhariwal and
  Kingma}{2019}]%
        {glow-face}
\bibfield{author}{\bibinfo{person}{Prafulla Dhariwal} {and}
  \bibinfo{person}{Durk Kingma}.} \bibinfo{year}{2019}\natexlab{}.
\newblock \bibinfo{booktitle}{\emph{Glow: Better Reversible Generative
  Models}}.
\newblock
\urldef\tempurl%
\url{https://openai.com/blog/glow/}
\showURL{%
\tempurl}


\bibitem[\protect\citeauthoryear{evgerher and sarmbruster}{evgerher and
  sarmbruster}{2019}]%
        {neo4j}
\bibfield{author}{\bibinfo{person}{evgerher} {and}
  \bibinfo{person}{sarmbruster}.} \bibinfo{year}{2019}\natexlab{}.
\newblock \bibinfo{booktitle}{\emph{RDKit-Neo4j project}}.
\newblock
\urldef\tempurl%
\url{https://github.com/rdkit/neo4j-rdkit}
\showURL{%
\tempurl}


\bibitem[\protect\citeauthoryear{G{\'o}mez-Bombarelli, Wei, Duvenaud,
  Hern{\'a}ndez-Lobato, S{\'a}nchez-Lengeling, Sheberla, Aguilera-Iparraguirre,
  Hirzel, Adams, and Aspuru-Guzik}{G{\'o}mez-Bombarelli et~al\mbox{.}}{2018}]%
        {gomez2018automatic}
\bibfield{author}{\bibinfo{person}{Rafael G{\'o}mez-Bombarelli},
  \bibinfo{person}{Jennifer~N Wei}, \bibinfo{person}{David Duvenaud},
  \bibinfo{person}{Jos{\'e}~Miguel Hern{\'a}ndez-Lobato},
  \bibinfo{person}{Benjam{\'\i}n S{\'a}nchez-Lengeling},
  \bibinfo{person}{Dennis Sheberla}, \bibinfo{person}{Jorge
  Aguilera-Iparraguirre}, \bibinfo{person}{Timothy~D Hirzel},
  \bibinfo{person}{Ryan~P Adams}, {and} \bibinfo{person}{Al{\'a}n
  Aspuru-Guzik}.} \bibinfo{year}{2018}\natexlab{}.
\newblock \showarticletitle{Automatic chemical design using a data-driven
  continuous representation of molecules}.
\newblock \bibinfo{journal}{\emph{ACS central science}} \bibinfo{volume}{4},
  \bibinfo{number}{2} (\bibinfo{year}{2018}), \bibinfo{pages}{268--276}.
\newblock


\bibitem[\protect\citeauthoryear{Jin, Barzilay, and Jaakkola}{Jin
  et~al\mbox{.}}{2018}]%
        {jin2018junction}
\bibfield{author}{\bibinfo{person}{Wengong Jin}, \bibinfo{person}{Regina
  Barzilay}, {and} \bibinfo{person}{Tommi Jaakkola}.}
  \bibinfo{year}{2018}\natexlab{}.
\newblock \showarticletitle{Junction tree variational autoencoder for molecular
  graph generation}.
\newblock \bibinfo{journal}{\emph{arXiv preprint arXiv:1802.04364}}
  (\bibinfo{year}{2018}).
\newblock


\bibitem[\protect\citeauthoryear{MODERN.DATA}{MODERN.DATA}{2019}]%
        {dashbio}
\bibfield{author}{\bibinfo{person}{MODERN.DATA}.}
  \bibinfo{year}{2019}\natexlab{}.
\newblock \bibinfo{booktitle}{\emph{Dash has gone full R}}.
\newblock
\urldef\tempurl%
\url{r-craft.org/r-news/dash-has-gone-full-r/}
\showURL{%
\tempurl}


\bibitem[\protect\citeauthoryear{Zang and Wang}{Zang and Wang}{2020}]%
        {moflow}
\bibfield{author}{\bibinfo{person}{Chengxi Zang} {and} \bibinfo{person}{Fei
  Wang}.} \bibinfo{year}{2020}\natexlab{}.
\newblock \showarticletitle{MoFlow: An Invertible Flow Model for Generating
  Molecular Graphs}. In \bibinfo{booktitle}{\emph{Proceedings of the 26th ACM
  SIGKDD International Conference on Knowledge Discovery \& Data Mining}}.
\newblock


\end{thebibliography}
}

\appendix

\end{document}